\documentclass[a4paper, 10pt, conference]{ieeeconf}     
\usepackage{FG2024}
\usepackage{graphicx}
\usepackage{subcaption}
\usepackage{array}
\usepackage{amsmath}
\usepackage{xcolor}
\usepackage{multirow}
\usepackage{float}
\usepackage{url}

\FGfinalcopy % *** Uncomment this line for the final submission

\IEEEoverridecommandlockouts                              % This command is only
\title{\LARGE \bf
CSTalk: Correlation Supervised Speech-driven 3D Emotional Facial Animation Generation
}

\author{\parbox{16cm}{\centering
    {\large Xiangyu Liang$^1$, Wenlin Zhuang$^1$, Tianyong Wang$^1$, Guangxing Geng$^2$, Guangyue Geng$^2$, Haifeng Xia$^1$, Siyu Xia$^1$}\\
    {\normalsize
    $^1$ School of Automation, Southeast University, Nanjing, China\\
    $^2$ Nanjing 8:8 Digital Technology Co., Ltd, Nanjing, China}}
    \thanks{This work was supported in part by Key Laboratory of Intelligent Processing Technology for Digital Music (Zhejiang Conservatory of Music), Ministry of Culture and Tourism under Grant 2023DMKLB003.}% <-this % stops a space
}

\usepackage{fancyhdr}
\begin{document}

\thispagestyle{empty}
\pagestyle{empty}

\maketitle
\thispagestyle{fancy}

\begin{abstract}
Speech-driven 3D facial animation technology has been developed for years, but its practical application still lacks expectations. The main challenges lie in data limitations, lip alignment, and the naturalness of facial expressions. Although lip alignment has seen many related studies, existing methods struggle to synthesize natural and realistic expressions, resulting in a mechanical and stiff appearance of facial animations. Even with some research extracting emotional features from speech, the randomness of facial movements limits the effective expression of emotions. To address this issue, this paper proposes a method called CSTalk (Correlation Supervised) that models the correlations among different regions of facial movements and supervises the training of the generative model to generate realistic expressions that conform to human facial motion patterns. To generate more intricate animations, we employ a rich set of control parameters based on the metahuman character model and capture a dataset for five different emotions. We train a generative network using an autoencoder structure and input an emotion embedding vector to achieve the generation of user-control expressions. Experimental results demonstrate that our method outperforms existing state-of-the-art methods. 

\end{abstract}

\section{INTRODUCTION}
The continuous development of computer graphics and artificial intelligence triggers significant advancements in speech-driven 3D facial animation generation. Early researchers generated facial animations based on rules, recognizing phoneme (the phonetic unit) sequences from audio and mapping them to corresponding lip movements~\cite{taylor2012dynamic}, and modeling collaborative pronunciation animations of adjacent phonemes by establishing a set of norms~\cite{xu2013practical} and designing dominance functions~\cite{massaro201212}. With the development of artificial intelligence~\cite{chi2018supervised}, Taylor et al.~\cite{taylor2017slidewindow} introduce deep learning methods~\cite{xia2020embedded} to learn collaborative pronunciation. However, recognition of phoneme sequences is a labor-intensive and low-precision task, so Karras et al.~\cite{karras2017nvidia} designed an end-to-end training method using TCN as the backbone, representing audio by traditional features such as MFCC or LPC. Subsequently, LSTM~\cite{pham2017bs}, transformer~\cite{fan2022faceformer} were also introduced to model audio signals.

Modeling a 3D virtual avatar encompasses two main approaches~\cite{shakir2022overview}: mesh-based models~\cite{cudeiro2019bs,karras2017nvidia,richard2021meshtalk} and parameterized models~\cite{bao2023tx,cudeiro2019bs,wang20213dbs}. Mesh-based models utilize the 3D Morphable Model~\cite{3dmm} to directly manipulate facial vertices, which allows intricate variations in facial expressions, while parameterized models employ a template-based framework to control facial movements by specific parameters. Among the parameterized models, blendshape models~\cite{terzopoulos1988bs} based on the ARKit 2 standard have been widely applied, which employ a set of predefined facial sub-motions as templates and generate diverse expressions by linearly combining the weights assigned to each sub-motion. This approach offers a high degree of controllability and generalization, enabling the reuse of animation parameters in different avatars. However, the linear combination of simple 52-dimensional data falls short in terms of achieving realistic and natural animations. In particular, accurately capturing subtle expressions in the upper facial region remains a challenge. Therefore, previous studies have focused mainly on aligning lip movements. Although some attempts have been made to incorporate emotional features into facial performances, such as Faceformer~\cite{fan2022faceformer} and Emotalk~\cite{peng2023emotalk}, these efforts focus on extracting emotional cues from audio, while neglecting the optimization of facial expression reconstruction.

In this study, we employ a parameterized model based on the MetaHuman character model proposed by Epic, which manipulates facial animation through 185 control rigs, with each rig corresponding to a group of facial muscles. By non-linearly deforming the facial vertices within their respective regions, the MetaHuman model exhibits potential in capturing intricate expressions. The remaining issue is to predict proper control rig curves from speech. Our study reveals that correlations exit between facial movements in different regions, stemming from both physical constraints of coordinated muscle control and habitual patterns. Specifically, during stress and speech pauses, regions such as the mouth, eyebrows, and cheeks tend to undergo simultaneous movements to convey expressive intent. Thus, we employ a transformer-based~\cite{vaswani2017attention} encoder to model the correlations. Using the model as a supervisor, we train a 3D facial animation generation model. This study contributes the following.
\begin{itemize}
    \item We propose CSTalk for speech-driven 3D emotional facial animation generation that enables to generate intricate expressions under different emotional states.
    \item We model the correlations among facial actions in specific emotions based on transformer encoders, serving as constraints to generate facial expressions that better mimic authentic human speech expressions.
    \item We first introduce a parameter model based on MetaHuman control rigs, which not only enables the synthesis of more natural and detailed animations comparable to mesh-based methods, but also makes the proposed workflow easier to integrate into industrial pipelines.
\end{itemize}
\begin{figure*}
    \centering
    \includegraphics[height=6cm,clip, trim=0cm 0cm 0cm 0.5cm]{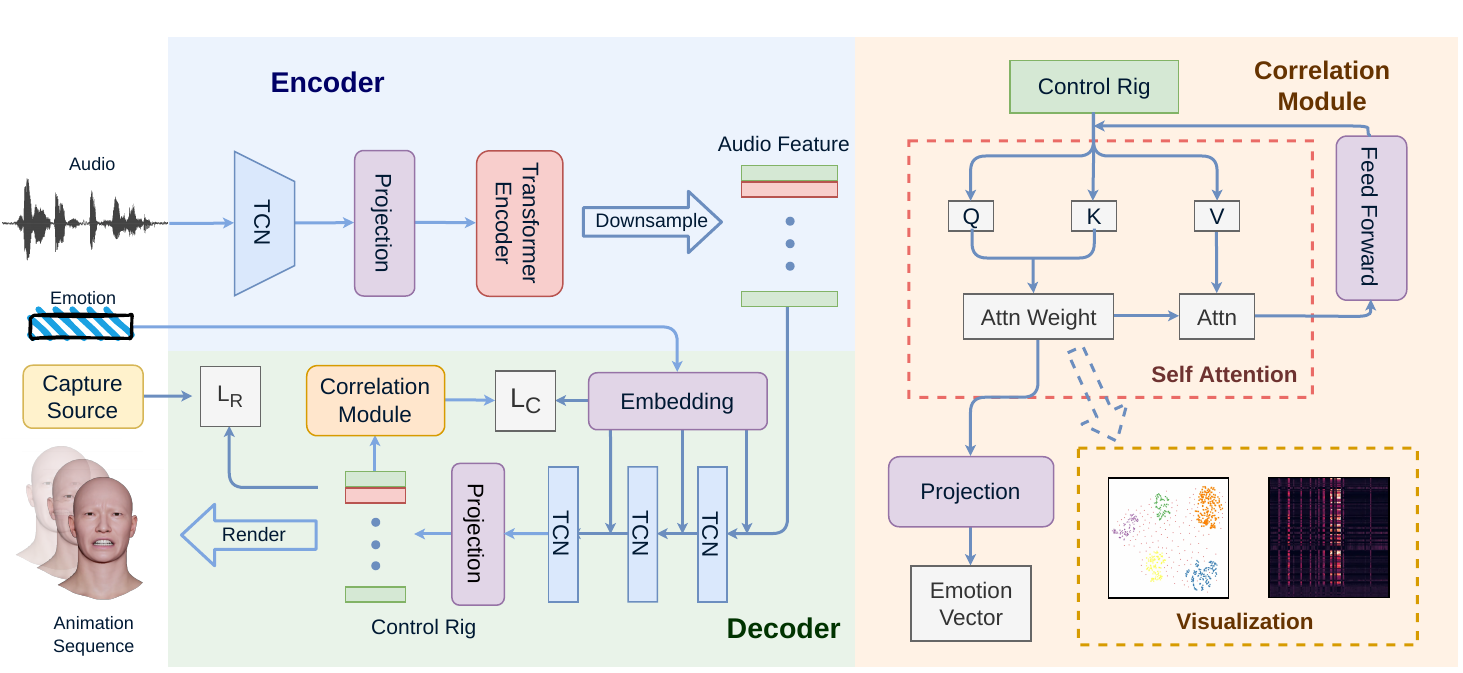}
    \caption{\textbf{The pipeline of CSTalk.} Our workflow consists of two stages. First, we input fixed length sequences of expressive speech animation data $S \in R^{R \times T}$ (where R is the amount of control rigs) to train the correlation module, along with the corresponding emotion labels. The module employs multiple transformer encoder layers to calculate attention weights, whose average is then fed into a linear layer to predict the emotion labels. In the second stage, an autoencoder takes audio data as input to generate a control rig sequence. In the decoder, the input of each TCN~\cite{bai2018empirical} layer is fused with the corresponding emotion embedding. The output is then passed through the pre-trained correlation module to predict the associated emotion.
}
    \label{fig:overall architecture}
\end{figure*}

\section{METHODS}\label{method}
\subsection{Animation Generation}\label{gen}
As shown in Fig.~\ref{fig:overall architecture}, we adopt an encoder-decoder model~\cite{hinton2006ae,xia2020hgnet} structure overall. The input data consists of audio information $X \in R^{Sam}$ via the Librosa library at a rate of 16kHz. To maximize data utilization, we segment the speech using sliding windows~\cite{taylor2017slidewindow} with large overlap. The encoder module utilizes the pre-trained Wav2vec 2.0 model proposed by Alexei~\cite{baevski2020wav2vec} et al. for audio feature extraction, which employs TCN and transformer encoder layers, demonstrating excellent performance in various speech recognition tasks. Compared to traditional algorithms such as MFCC and LPC, Wav2vec 2.0 is more suitable for extracting phonetic features, representing units of pronunciation, which makes it well suited for this task.

Previous works have also used Wav2vec 2.0 for audio encoding~\cite{fan2022faceformer,peng2023emotalk}, with a decoder based on intuitive use of a transformer decoder. However, this structure requires autoregressive output, similar to machine translation tasks, where each frame output depends on the decoding of all audio features. Actually, facial animation exhibits minimal dependencies on distant time frames, especially in lip movements. Except for the cooperative articulation effect of adjacent phonemes, most lip movements are largely independent. Predicting expressions for long input speech would introduce unnecessary high time complexity and increase the difficulty of model convergence. Therefore, we choose to construct the decoding module using TCN networks and control the influence of each frame on its neighborhood by designing the receptive field. Furthermore, each layer is fused with an emotion embedding $e$ to enable emotion control. Finally, a linear layer output animation parameters curve $y \in R^{T\times R}$, including T frames of R control rig data. The loss function for this phase is designed as follows:
\begin{equation}
\setlength{\jot}{10pt}
\begin{aligned}
    L_R &= \frac{1}{N}\sum_{i=1}^N||y_i-\hat{y_i}||_2^2 \\
       &+ \frac{1}{N-1}\sum_{i=1}^{N-1}||(y_{i+1}-y_i)-(\hat{y}_{i+1}-\hat{y}_i)||_2^2
\end{aligned}
\end{equation}
where $N$ denotes the number of samples in the training dataset, $y$ and $\hat{y}$ denote the groudtruth of control rig curves and the prediction, respectively. 
The latter part indicates the differential between adjacent frames of curves, which is designed for smooth generation.

\subsection{Correlation Modeling}\label{cor}
Facial muscles typically coordinate to control facial movements. For example, a smile usually involves the simultaneous activation of several muscles such as the zygomaticus major~\cite{facemuscle}, the buccinator, and the risorius. This coordination is reflected in the control rigs near the lips, such as mouth\_dimple, mouth\_cornerpull, and mouth\_sharpcornerpull. If these parameters vary independently, the generated facial movement will appear weird. The constraints interpreted above in the neighboring facial regions can be categorized as physical correlations. Furthermore, some different regions synchronize their movements to express the same category of information. For example, when experiencing anger, contraction of the corrugator muscle and the corrugator supercilii muscles causes the brow to furrow. And during speech, the contraction of the levator labii superioris alaeque nasi muscle raises the upper lip and nostril. The corresponding control parameters also exhibit synchronized variations. 

Intuitively, we assume that there exists a correlation among different control parameters in common facial expressions. The correlation varies changes with different emotions, but there is a set of common constraints followed during speech, primarily composed of physical constraints between adjacent muscles. Guided by this intuition, we train a correlation model for each emotion. Considering the excellent ability of transformers to model correlation, we design a correlation module based on the transformer encoder. In order to enable the module to calculate the correlation of different rigs, we select the control rig as the axis of the sequence for the input vector, represented as $S \in R^{R \times T}$. The self-attention formula is as follows:
\begin{equation}
    Attn(Q,K,V)=\frac{Softmax(Q\times K^T)}{\sqrt{d_{model}}}\times V
\end{equation}
where $QK^T$ plays the role of attention weight, representing the influence among different positions in the sequence. After multiplying by V representing the sequence itself, the feature vector at each position of the sequence is updated with information from others. While what we need is the correlation of different positions, and considering that the computational form of $QK^T$ is consistent with cosine similarity, we adopt it as the output of the module. Here, we sum the scores of each layer to obtain the correlation features. We utilize two fully connected layers to predict emotion vectors from correlation features.
The loss function in this phase is as follows.
\begin{equation}
    L_C=-\frac{1}{N}\sum_{i=1}^N\sum_{c=1}^Me_{ic}log(\hat{e}_{ic})
\end{equation}
where $M$ denotes the number of classifications, $e_{ic}$ denotes whether the real label of the $i^{th}$ sample is the $c^{th}$ classification, and $\hat{e}_{ic}$ denotes the predicted probability. The training dataset also includes random data in the same quantity as other emotional data, helping the model to discriminate valid facial animation curves.
\subsection{Correlation Supervision}
We utilize the pre-trained correlation model to supervise the training of the generation module in Section~\ref{gen}, enforcing constraints among expression coefficients under specific emotional contexts. When feeding predicted animation curves into a correlation module, the resulting predictions are compared with the emotional vectors as a loss. Consequently, the loss function for expression generation becomes:
\begin{equation}
    L_G=L_R+L_C
\end{equation}
Note that $L_C$ shares the same formula as that in Section \ref{cor}, but it trains the generation model instead of correlation.
\begin{figure}[t]
    \centering
    \begin{subfigure}{0.5\textwidth}
        \centering
        \includegraphics[width=0.8\linewidth,clip,trim=0cm 1.2cm 0cm 0cm]{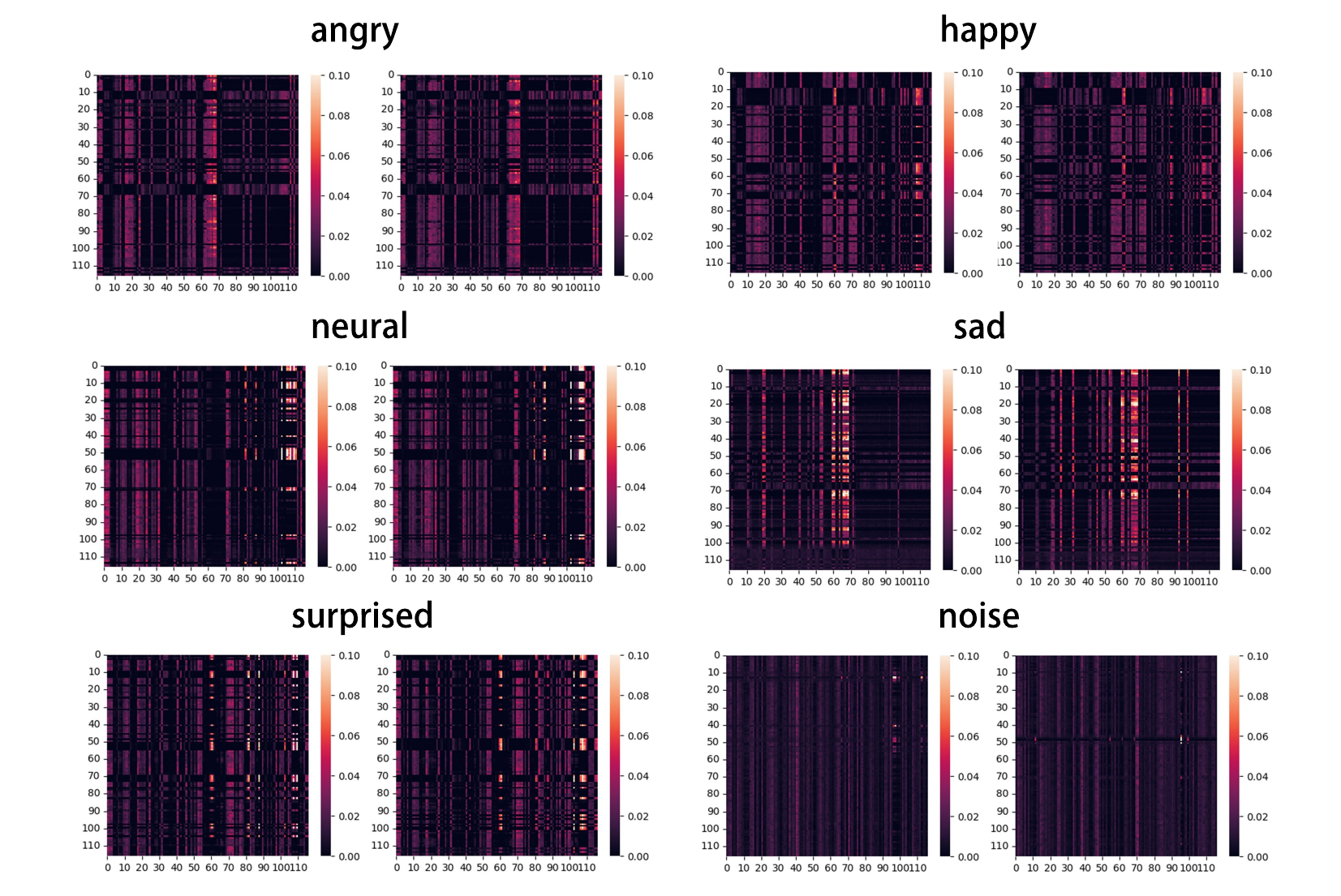}
        \caption{}
        \label{fig:a}
    \end{subfigure}
    \hspace{0.1cm}
    \begin{subfigure}{0.3\textwidth}
        \centering
        \includegraphics[width=0.8\linewidth,clip,trim=0cm 1.3cm 0cm 2cm]{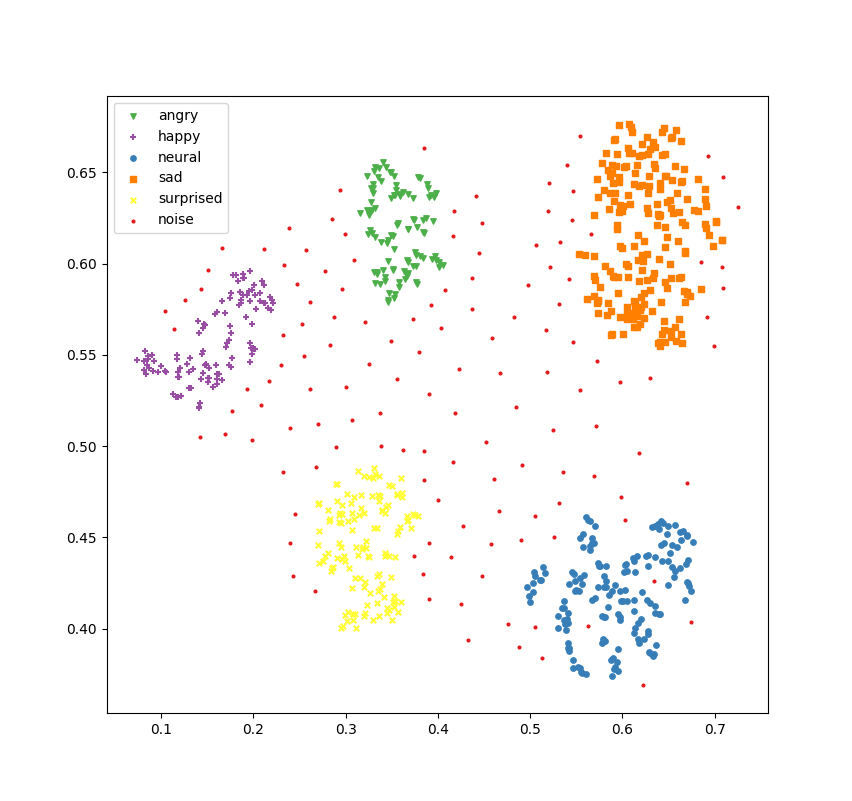}
        \caption{}
        \label{fig:b}
    \end{subfigure}
    \caption{\textbf{Visualization of score matrices.} (a) shows heatmaps of attention weights of different emotions, including four pieces of animation data for each emotion. (b) shows the visualization of the weight data that is subjected to dimensionality reduction using t-SNE.}
    \label{fig:correlation_visualize}
\end{figure}
\section{EXPERIMENT}
\subsection{Dataset}
Current open-source facial animation datasets such as VOCA~\cite{cudeiro2019bs} only include mesh data. Therefore, we conducted our own data collection using the Live Link Face application on an iPhone and utilized the MetaHuman Animator plugin in Unreal Engine 5 to compute the control rig data. The dataset comprises five emotions: neutral, angry, sad, surprised, and happy, with 100 samples for each emotion. Each sample ranges from 4 to 9 seconds in duration.
\subsection{Model Training}\label{train}

The correlation module is trained separately.  We excluded certain control rigs that remained constant throughout all animations, leaving 116 remaining. Each rig vector contained 96 frames of continuous data (at a frame rate of 30 fps). The batch size and the learning rate are set to 64 and 1e-5, respectively. 

During training the generation model, we sample the speech audio at a rate of 16 kHz using the Librosa library, employ a sliding window of 96 frames (corresponding to 51200 audio samples) and a stride of 5 frames (corresponding to 2667 audio samples). The processed audio data is fed into the pre-trained Wav2vec 2.0 model, which includes 7 TCN layers, a projection layer, and 12 layers of transformer encoder layers, outputting a set of audio features $A \in R^{159\times 116}$. The decoder then downsamples the features to 96 frames and passes through the TCN layers. The batch size and the learning rate are set to 64 and 1e-6, respectively. In this phase of training, the parameters of the TCN layer in the encoder and the correlation supervision module are frozen.

\subsection{Result}\label{result}
Attention weights after training are visualized in Fig.~\ref{fig:correlation_visualize}. (a) represents the attention weights data as the correlation matrices for different control rigs. Shows that, under specific emotions, the correlation matrices for different data exhibit similar distributions. Moreover, compared to the results obtained with random noise input, the correlation matrices for all expression data share commonalities, which indicates that the correlation module can to some extent reflect the shared relationships among different facial regions and the specificity of these relationships under different emotions. Fig.~\ref{fig:correlation_visualize} (b) shows the visualization of the attention weights using t-SNE, which reveals clearer clustering patterns.

\begin{figure}
    \centering
    \includegraphics[width=0.8\linewidth]{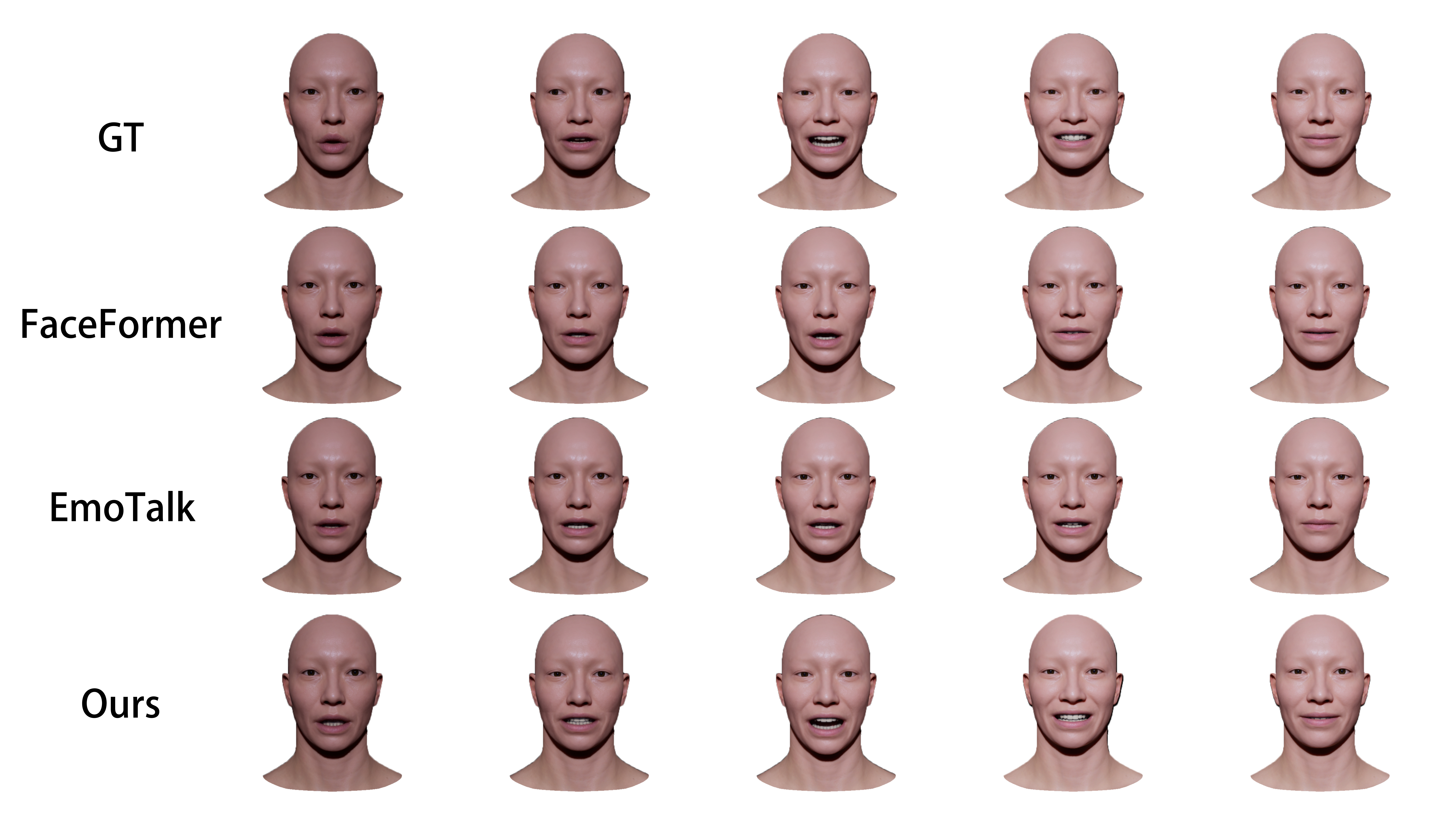}
    \caption{\textbf{Qualitative comparison of the rendered animation frames.} We compare our results with the SOTA methods in "happy" emotion, based on the same avatar.}
    \label{fig:contrast}
\end{figure}

To compare with other methods, we train their models with our dataset and render the predicted animations on the same avatar. The animated results generated are shown in Fig.~\ref{fig:contrast}. Our method achieves high performance both in aligning lip movement with speech and in natural expressions that fit specific emotions. To evalute the lip synchronization between the predicted face animation and groundtruth, we adopt the lip vertex error (LVE) employed in Meshtalk~\cite{richard2021meshtalk}, which is defined as the maximum l2 error of all lip vertices for each frame. We report the average of l2 error in all frames in the test set generated by Faceformer~\cite{fan2022faceformer}, Emotalk~\cite{peng2023emotalk} and our method. Since LVE is not able to evalute the emotion quality of predictions, we also follow Emotalk to use the emotional vertex error (EVE), which is similar to LVE, while it concentrates on the eye and forehead regions. The results are shown in Tab.~\ref{tab:metrics}, indicating an excellent performance of our model.

\begin{figure}
    \centering
    \includegraphics[width=0.8\linewidth]{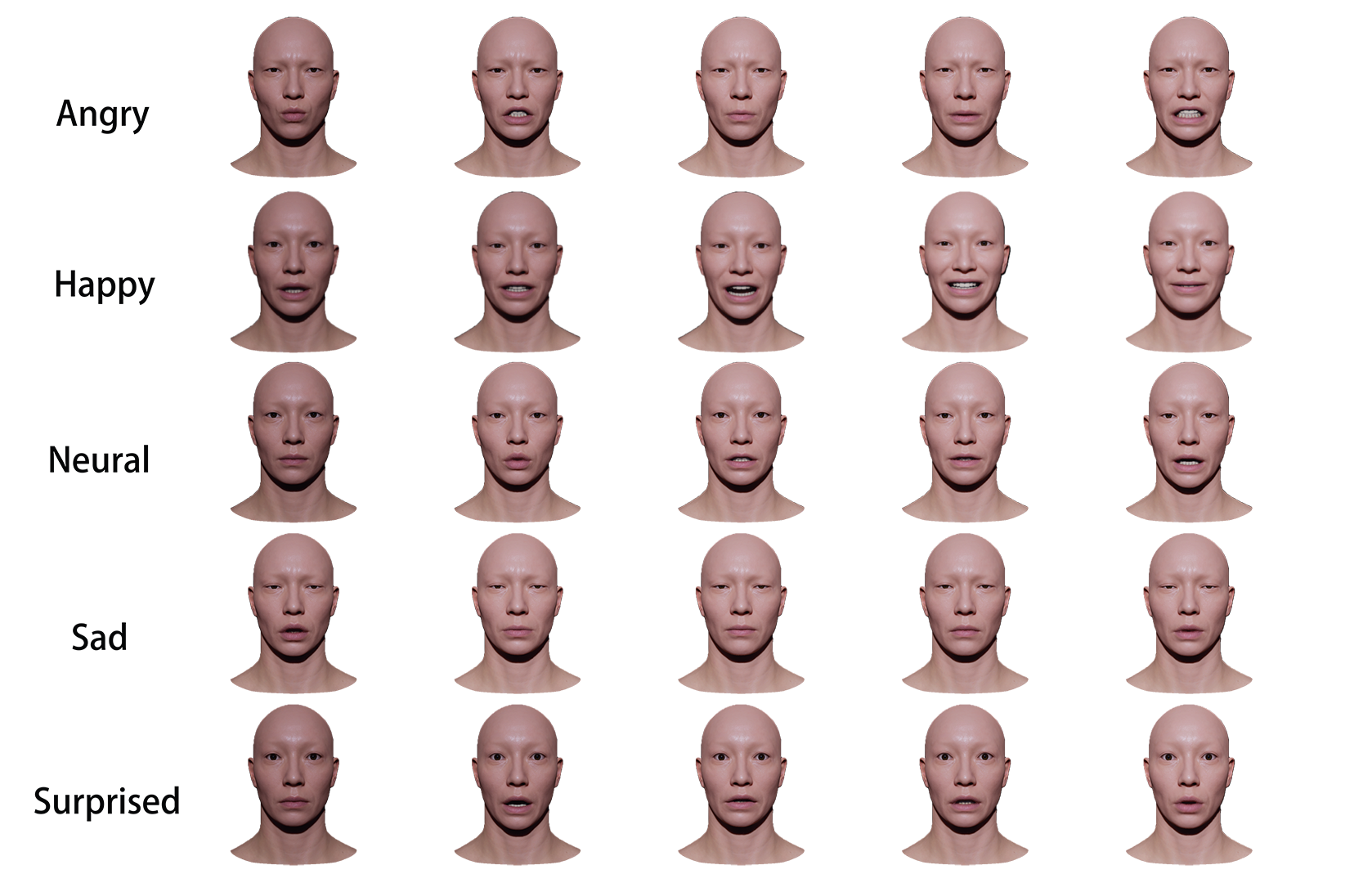}
    \caption{\textbf{Rendered results.} Some frames of predicted animations in 5 emotions.}
    \label{fig:rendered_result}
\end{figure}

\begin{table}[ht]
\caption{\textbf{Qualitative metrics compared to other methods.} We evaluated the experimental results using LVE and EVE, and achieved the highest performance compared to two state-of-the-art methods.}
\centering
\renewcommand\arraystretch{1.2}
\begin{tabular}{@{\hspace{15pt}}l@{\hspace{15pt}}|@{\hspace{15pt}}c@{\hspace{15pt}}|@{\hspace{10pt}}c@{\hspace{15pt}}}
\hline
Methods       & LVE(mm)↓       & EVE(mm)↓       \\ \hline
FaceFormer~\cite{fan2022faceformer}    & 3.511          & 3.069          \\
EmoTalk~\cite{peng2023emotalk}       & 2.954          & 2.875          \\
\textbf{Ours} & \textbf{2.538} & \textbf{2.084} \\ \hline
\end{tabular}
\label{tab:metrics}
\end{table}

\subsection{Ablation Experiment}
In the ablation experiment, we examine the effects of the following parts on the results: the introduction and output of the correlation module, and the selection of the audio encoder. As shown in Tab.~\ref{tab:ablation}, both errors increase without the correlation module, indicating a poor quality of lip sync and facial expressions. For training the correlation module, predictions perform worse when selecting logits of the transformer encoder as output instead of attention weights. 
But it should be noticed that EVE exhibits a relatively low level. According to our analysis, when directly outputting logits, the model extracts features belonging to certain types of emotion from the animation input, resulting in generated animations that are more in line with emotions, which is more like an emotion analysis module. 
Finally, we also tested the selection of audio encoders and found that using wav2vec 2.0 as an encoder can achieve significantly better results compared to traditional MFCC.
\begin{table}[H]
\caption{\textbf{Ablation experiments.} We show quantitative results in different cases.}
\centering
\renewcommand\arraystretch{1.2}
\begin{tabular}{@{\hspace{10pt}}l@{\hspace{10pt}}|@{\hspace{10pt}}c@{\hspace{10pt}}|@{\hspace{10pt}}c@{\hspace{10pt}}}
\hline
Methods       & LVE(mm)↓       & EVE(mm)↓       \\ \hline
w/o Correlation Module    & 3.502          & 3.216          \\

Logits output & 3.348 &3.966 \\
MFCC Encoder    & 5.178          & 3.966          \\
\textbf{Ours} & \textbf{2.538} & \textbf{2.084} \\ \hline
\end{tabular}
\label{tab:ablation}
\end{table}
\section{CONCLUSION}
This paper proposes a speech-driven 3D emotional facial animation generation network CSTalk, capable of generating animations with aligned lip movement and realistic expressions. And we first introduce MetaHuman-based facial control rig models, enabling direct collaboration with artists and application in industrial pipelines. The generated animation parameters are identity-free and can be reused for any metahuman avatar. Additionally, we reveal the existence of correlations between different regions of facial movement and model these correlations under different emotions and use them to help train the network to generate expressions that adhere better to facial motion patterns. Our method outperforms existing approaches in terms of results.

\addtolength{\textheight}{-3cm}   
\newpage
{\small
\bibliographystyle{ieee}
\bibliography{egbib}
}

\end{document}